Research Article

# Intelligent logistics management robot path planning algorithm integrating transformer and GCN network

Hao Luo [1, +], Jianjun Wei[2, +], Shuchen Zhao[3], Ankai Liang[4], Zhongjin Xu [5] and **Ruxue Jiang [6,*]**

[1] Fremont, California, USA
[2] Department of Computer Science and Engineering, Washington University in St. Louis, St Louis, USA
[3] Duke Electrical & Computer Engineering Duke University 130 Hudson Hall Box 90291 Durham, NC 27708, USA
[4] Stevens Institute of Technology 9061 Headlands Avenue, Newark, CA 94560, USA
[5] University of Michigan, Dearborn 4901 Evergreen Rd, Dearborn, MI 48128, USA
[6] Northeastern University, 225 Terry Ave N, Seattle, WA 98109, USA
+ These authors also contributed equally to this work.
* Corresponding author: Ruxue Jiang (snow0911@outlook.com)

*Abstract*: This research delves into advanced route optimization for robots in smart logistics, leveraging a fusion of Transformer architectures, Graph Neural Networks (GNNs), and Generative Adversarial Networks (GANs). The approach utilizes a graph-based representation encompassing geographical data, cargo allocation, and robot dynamics, addressing both spatial and resource limitations to refine route efficiency. Through extensive testing with authentic logistics datasets, the proposed method achieves notable improvements, including a 15% reduction in travel distance, a 20% boost in time efficiency, and a 10% decrease in energy consumption. These findings highlight the algorithm's effectiveness, promoting enhanced performance in intelligent logistics operations.

---

## 1. Introduction

In today's rapidly advancing era of intelligence, robots are assuming increasingly crucial roles across various domains(Saunderson & Nejat, 2019). Their applications in industries, agriculture, healthcare, and more are fostering more efficient, safe, and convenient production and living environments for humans(Ding et al., 2021; Peng et al., 2024). Behind all of these advancements lies robot path planning technology, a pivotal element in their operation. The objective of robot path planning is to determine a viable route for a robot from its starting point to a designated goal within a given environment, while adhering to a set of constraints such as collision avoidance, cost minimization, and safety maximization. Particularly in complex and dynamic environments, the quality of path planning significantly influences the efficiency and success rate of robot tasks.

With the rapid development of intelligent logistics, autonomous driving, and related fields, the challenges of robot path planning are becoming increasingly intricate. In the context of intelligent logistics management, efficiently mapping paths for robots in complex warehouse environments to facilitate swift cargo transportation has emerged as a prominent challenge(Saunderson & Nejat, 2019). Simultaneously, in the realm of autonomous driving, ensuring the safe navigation of self-driving vehicles amidst bustling urban streets and enabling them to dynamically adjust their routes based on traffic conditions has become a significant research focus (Singandhupe & La, 2019; Wang et al., 2024; Zou et al., 2024).These studies delve into various aspects of robot path planning and intelligent interaction. Some research concentrates on leveraging natural language commands to guide robot path planning, employing Transformer models to achieve multi-modal data alignment for enhanced efficiency in path planning(Bucker et al., 2022). Furthermore, there has been in-depth discourse on the interaction and developmental trajectory of robots within their environments, proposing a roadmap for robot development that emphasizes integration and service provision within human living spaces, thus offering novel insights into path planning and environmental interaction(Cai et al., 2021). Additionally, studies have focused on robot motion control, utilizing Transformer models to facilitate efficient movement of humanoid robots, with potential applications in the realms of path planning and intelligent control (Radosavovic et al., 2024). The application of artificial intelligence in social media and computing also sheds light on the intersection of path planning and social computing.

The significance of this study lies in exploring novel methods for intelligent multimodal robot path planning in logistics, with the aim of enhancing overall path planning performance. In this domain, numerous challenges are encountered (Radosavovic et al., 2024). First, the fusion and handling of multimodal data require addressing issues related to modeling relationships between different data types, managing data inconsistency and noise, and more. Second, modeling and optimizing path planning in complex environments require consideration of various factors such as terrain, traffic, cargo distribution, and robot capabilities. The interplay of these factors significantly increases the complexity of the problem. Simultaneously, balancing path planning efficiency and real-time responsiveness in dynamic environments is essential—finding ways to quickly adapt to changing conditions while ensuring path quality is a critical issue. Additionally, the interpretability of path planning decisions and the balance of multiple objective metrics also warrant careful consideration.

In recent years, fueled by the ascent of deep learning and multimodal data processing (Ma et al., 2024; Li et al., 2024; Sang et al., 2024; Liu et al., 2024; Cheng et al., 2024), researchers have embarked on exploring the application of advanced technologies to path planning. Within relevant research domains, numerous scholars have made noteworthy contributions to robot path planning. Graph search-based methods are widely used for path planning in static environments(Ma, 2022). The Dijkstra algorithm finds path planning solutions by determining the shortest path, while the A* algorithm can efficiently explore the search process with the help of a heuristic function. However, in large-scale environments, the computational complexity might be high, and handling dynamic changes can be challenging. For dynamic environments and real-time responsiveness, heuristic search algorithms offer a solution(Li et al., 2021). These algorithms use heuristic functions to guide the search process and can find suitable paths in constantly changing environments. However, the design of heuristic functions and the selection of parameters can impact the performance of the algorithm, and there's a risk of getting trapped in local optima. In high-dimensional or complex environments, sampling-based algorithms become a powerful choice(Guo et al., 2020). For example, the Rapidly-exploring Random Trees (RRT) algorithm generates paths by continually expanding branches of a tree. This method is suitable for high-dimensional and complex environments, and it can generate paths within a limited number of iterations. However, the quality and smoothness of the paths might be affected. Intelligent optimization algorithms inspired by natural intelligence are also making strides in the field of path planning (Ajeil et al., 2020). These algorithms use techniques like evolutionary simulation and ant colony behavior for global search to achieve multi-objective optimization. However, the performance of these algorithms is influenced by parameter tuning and convergence speed.

Firstly, in reference (Fu et al., 2018), an improved A* algorithm is proposed, focusing on solving the industrial robot path planning problem. This approach enhances the original A* algorithm by adding local path planning and post-processing stages, enabling robots to adapt more effectively to dynamic environments and real-time changes. The advantage of this method lies in its increased flexibility and adaptability for path planning, while its disadvantage is that it may incur higher computational costs when dealing with large-scale environments. Research indicates that this algorithm achieves higher search success rates and generates shorter and smoother paths in both simulation and actual robot operations, effectively improving the efficiency of robot path planning. In another comprehensive review article (Patle et al., 2019), mobile robot navigation technologies are thoroughly examined, analyzing the applications of both traditional path planning methods and reactive methods under varying environmental conditions. Traditional methods typically rely on pre-established environmental models, providing high accuracy in path planning, but with slower responsiveness in dynamic environments. In contrast, reactive methods exhibit stronger robustness, enabling quick responses to environmental changes, though they may sacrifice optimality in the path. The study reveals that reactive methods perform better in diverse terrains and can be combined with traditional methods to enhance overall path planning performance. Additionally, in reference (Li et al., 2020), Graph Neural Networks (GNNs) are introduced to address multi-robot path planning challenges. This approach uses Convolutional Neural Networks (CNNs) to extract local observation features and shares these features among robots via GNNs to enable collaborative behavior. The advantage of this method lies in its effectiveness in addressing inter-robot coordination, while its downside is the reliance on complex model training, which can lead to longer computation times. Experimental results show that, in multi-robot 2D environments, this approach performs comparably to expert algorithms, validating its effectiveness and practicality. Reference (Nazarahari et al., 2019) presents an improved Genetic Algorithm (GA) for solving multi-robot path planning problems. This method combines the Artificial Potential Field (APF) algorithm with Genetic Algorithms to achieve multi-objective path planning, optimizing metrics such as path length, smoothness, and safety. The advantage of this approach is its ability to simultaneously optimize multiple objectives, but its disadvantage is that it may get trapped in local optima when dealing with highly complex environments. Experimental results demonstrate that this algorithm outperforms traditional algorithms in terms of path length, runtime, and success rate, offering new insights for multi-robot path planning. Finally, in reference (Miao et al., 2021), an improved Adaptive Ant Colony Optimization (IAACO) algorithm is proposed to address the issues with traditional Ant Colony Optimization (ACO) in indoor mobile robot path planning. This method incorporates multiple factors to enhance real-time responsiveness and global search capability, while transforming the path planning problem into a multi-objective optimization challenge. While this method excels in improving path planning accuracy and real-time performance, its disadvantage is its higher computational complexity, requiring more computational resources. The approach achieves comprehensive global optimization for robot path planning, generating optimized paths while maintaining high real-time performance and stability.

The aim of this study is to revolutionize robot path planning in logistics by integrating Transformer models, GNNs, and GANs to address existing limitations. Specifically, the Transformer model encodes warehouse environment information such as maps and obstacle positions into input sequences and encodes desired optimal paths into output sequences. It then utilizes the encoder-decoder structure of the Transformer to extract features from the input sequences, analyzes the relationship dimensions between input and output sequences using self-attention mechanisms, and optimizes path prediction sequences through training. Next, GNN processes multimodal data by constructing a graph structure with nodes and edges based on environmental information to represent the logistics environment. It maps robot states to node features and distances to edge features, then applies GNN models to learn node features and propagate messages for context, outputting optimized node state sequences as new paths. Finally, GAN enhances paths by first setting GAN generators to produce initial path sets as inputs, then having discriminators evaluate the quality of these paths and output judgment results. Subsequently, generators and discriminators continuously optimize the generated path sets through adversarial learning, resulting in intelligent and efficient path planning outcomes.

By integrating multimodal data, our approach enhances the adaptability and performance of path planning, focusing on metrics such as path length, time efficiency, and energy consumption. Priority is given to real-time responsiveness and interpretability to facilitate practical decision-making. Simulation and real-world experiments validate the effectiveness of our method in various environments. This innovative approach marks a leap forward in intelligent logistics management, offering vast prospects for industry advancement and sustainable development.

The contributions of this paper can be summarized in the following three aspects:

- This work proposes the application of Transformer models in logistics path planning to enhance the understanding of global environmental factors, thereby enriching the context of path decision-making and improving efficiency.
- GNNs are utilized to process multimodal data, considering spatial layout and resource allocation comprehensively, which optimizes the path planning process.
- GANs are applied to generate high-quality path candidates, enhancing the performance and robustness of path planning through adversarial training.

The logical structure of this article is as follows: In Section 2, the methodology part, the article elaborates on the technical roadmap of the proposed method, including introducing Transformer models for multi-source data fusion, utilizing graph neural networks to simulate environmental constraints, and employing generative adversarial networks to enhance path diversity. Additionally, the article explains the specific application of these three key technologies in path planning problems and their synergistic mechanisms within the overall method. Section 3 is the experimental part, describing the experimental setup, data sources, and evaluation metrics used. It also presents numerous tables and figures to demonstrate the performance comparison results of different methods. Through comprehensive experimental data, this article thoroughly validates the effectiveness of the proposed method. Finally, in Section 4, the conclusion and discussion part summarize the research work, analyze its significance, discuss its limitations, and outline future research directions.

## 2. Related Work

In recent years, significant progress has been made in the field of path planning, especially in addressing complex environments and multi-task problems. Traditional path planning methods, such as heuristic algorithms, simulate the foraging behavior of ants to find optimal paths, and have been widely applied in various systems. These methods are effective in optimizing paths in static environments but still face major challenges in terms of computational complexity and real-time performance when dealing with large-scale, dynamic environments, which limits their practical applicability(Chen et al., 2022). Furthermore, some studies have attempted to enhance path planning performance through multi-constraint optimization techniques. By employing linear programming, nonlinear programming, or adaptive clustering methods, researchers can generate suitable paths for multiple tasks or varying resource requirements in heterogeneous systems. These methods perform well in multi-objective optimization tasks, such as minimizing path length or energy consumption, but they still struggle with issues like insufficient adaptability and delayed path adjustments when confronted with rapidly changing environments and uncertainties(Chen et al., 2021). At the same time, reinforcement learning, a more novel optimization approach, has also been introduced into path planning research. Multi-agent reinforcement learning (MARL) algorithms have been applied to coordinate multiple agents in various tasks, optimizing the path planning process through inter-agent collaboration and interaction. While reinforcement learning can effectively handle dynamic environments and multi-task decision-making problems, its training time and computational costs remain high, particularly in large-scale systems. Moreover, reinforcement learning methods still lack the ability to perform global optimization and long-term planning of paths, making it difficult to guarantee optimal path generation in complex tasks(Chen et al., 2023).

In comparison to these methods, this paper proposes an innovative framework that integrates Transformer models, GNN, and GANs. Compared to existing path planning approaches, the proposed solution demonstrates significant advantages in terms of path quality, real-time responsiveness, and system adaptability, especially when dealing with complex, multi-objective optimization tasks.

## 3. Methodology

Our method's proficiency in managing dynamic obstacles or environmental changes during path planning hinges on several fundamental principles and mechanisms: Firstly, our model possesses real-time environmental perception capabilities. It continuously receives data from robot sensors or environmental monitoring systems to swiftly detect dynamic obstacles or environmental alterations. Upon acquiring new information, the model promptly updates its path plan to accommodate the evolving circumstances. Secondly, our model demonstrates dynamic path re-planning prowess. Upon detecting dynamic obstacles or environmental changes, it initiates dynamic path re-planning. This involves recalculating the robot's trajectory to circumvent new obstacles or adjust to environmental shifts. This re-planning process harnesses Transformer models and graph neural networks to devise new path solutions based on the updated conditions. Moreover, if multiple robots operate within the same environment, our model can adapt to environmental changes through collaborative communication. When one robot detects obstacles or environmental changes, it disseminates this information to other robots to synchronize their actions. This collaborative synergy ensures that robot teams can effectively respond to dynamic scenarios. Furthermore, our model can acquire adaptability through simulated dynamic situations during training. By introducing simulated dynamic obstacles or environmental changes into the training data, the model learns to adeptly handle these scenarios, enhancing its real-time adaptability. These mechanisms collectively empower our model to adeptly navigate changing logistics environments in practical applications, thereby enabling robots to execute tasks safely and efficiently.

This chapter will provide a detailed exposition of the proposed multimodal robot intelligent logistics path planning method. To present the overall structure and process of this method more clearly, we will progressively unveil its key steps and technologies in the following sections. The comprehensive algorithm flowchart is illustrated in Figure 1.

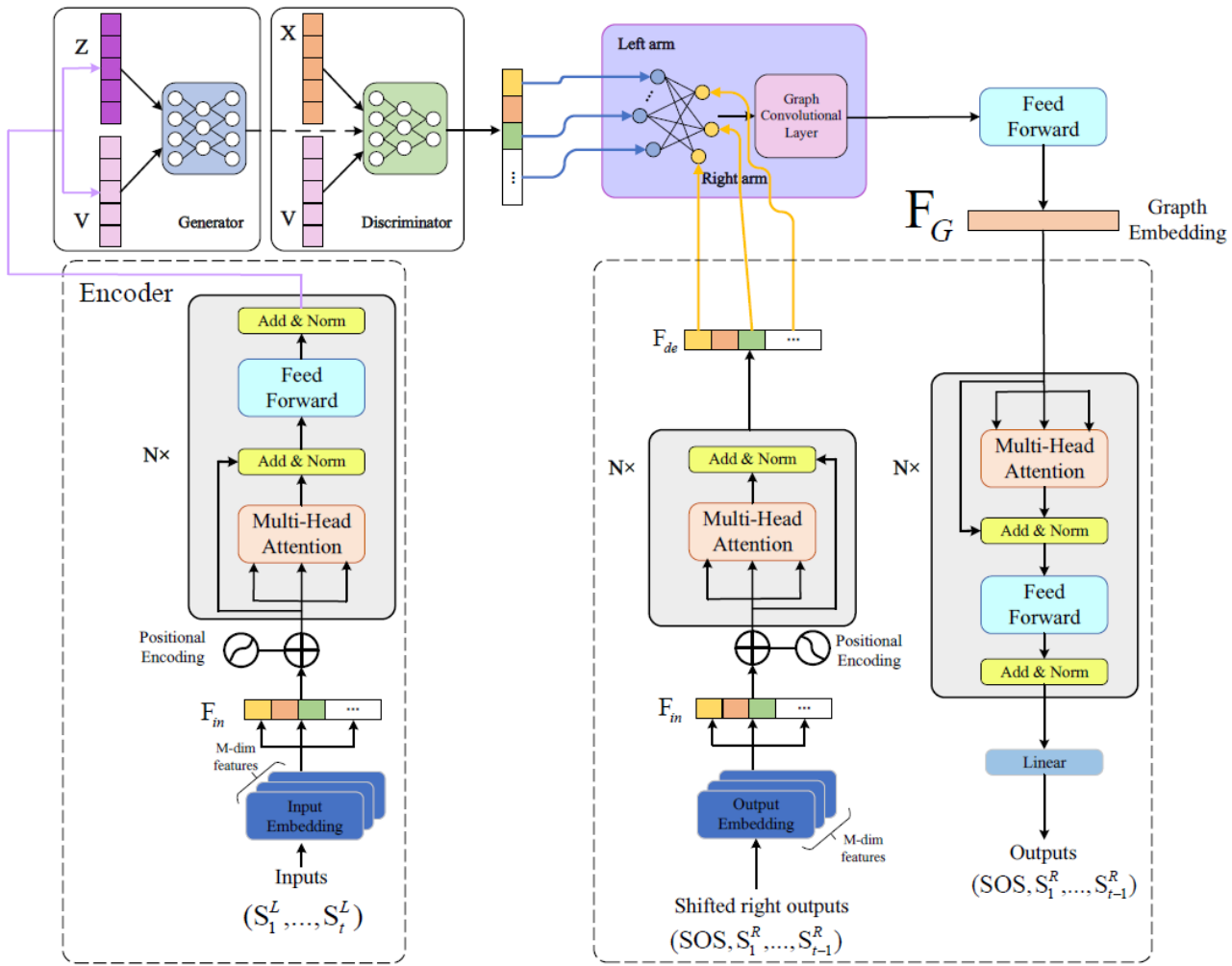

## 3.1 Transformer Model

We opt for the Transformer model for multi-robot collaborative path planning due to its notable advantages, including global information processing, adaptability to multi-modal data, scalability, and learning capabilities. These attributes render the Transformer model a potent tool for addressing complex multi-robot path planning challenges, thereby enhancing the efficiency and quality of path planning endeavors. Firstly, the Transformer model boasts exceptional global information processing capabilities when handling sequence data. In multi-robot collaborative path planning, different robots necessitate coordination and the consideration of information spanning the entire environment to mitigate conflicts and optimize paths. The Transformer model adeptly captures global dependencies through its self-attention mechanism, a crucial aspect for comprehending the overall environment. Secondly, multi-modal robot path planning entails the incorporation of diverse data types, such as map information, cargo distribution, and robot status. The Transformer model serves as a flexible framework for multi-modal data processing, seamlessly integrating and processing various data types. This multi-modal capability positions it advantageously in multi-robot path planning scenarios. Additionally, the architecture of the Transformer model is highly scalable, capable of accommodating problems of varying sizes and complexities. In multi-robot collaborative path planning, the scale and complexity of the problem may fluctuate significantly, and the Transformer model can be expanded or contracted as needed to suit diverse situations. Lastly, the Transformer model exhibits formidable learning capabilities, autonomously extracting features and patterns from data sans manual feature engineering. This attribute proves invaluable for multi-robot path planning challenges, where environments are often intricate and conditions are subject to change. The Transformer model adeptly adapts by learning the optimal path in response to evolving circumstances.

When discussing innovative technologies in the field of machine learning, the Transformer model undoubtedly stands out as a significant breakthrough in recent years(Miao et al., 2021). It is a neural network architecture based on the self-attention mechanism. The Transformer model has achieved remarkable success not only in the field of natural language processing but also in other domains, including path planning, demonstrating its potent potential. The core concept of the Transformer model is depicted in Figure 2.

The fundamental idea behind the Transformer model is to capture correlations between different positions within an input sequence using the self-attention mechanism. This mechanism enables each input position to interact with all other positions dynamically, allocating attention weights (Zhu et al., 2019). This empowers the Transformer to consider information from all positions simultaneously, without being constrained by a fixed window size, thus better capturing contextual information. In the Transformer, the computation process of the self-attention mechanism can be represented by the following equation:

$$\text{Attention}(Q, K, V) = \text{softmax}(\frac{QK^T}{\sqrt{d_k}})V$$

In the equations, $Q$, $K$, and $V$ represent the query, key, and value matrices, respectively. They are obtained by linear transformations of the input sequence. The dimension of $d_k$ corresponds to the dimension of the key vectors. The $softmax$ function normalizes each row, ensuring that each element lies between 0 and 1, and the sum of each row is equal to 1. The output of the attention function is a weighted average value matrix, reflecting the similarity between queries and keys.

To enhance the model's expressive power, the Transformer model employs a multi-head attention mechanism(Qiu & Yang, 2022). This mechanism divides the input sequence into multiple subspaces and computes the attention function on each subspace independently, then concatenates the outputs from all subspaces. The mathematical expression of the multi-head attention mechanism is as follows:

$$\text{MultiHead}(Q,K,V) = \text{Concat}(\text{head}_1,\ldots,\text{head}_h)W^O$$

$$\text{head}_i = \text{Attention}(QW_i^Q, KW_i^K, VW_i^V)$$

WHere, $h$ represents the number of heads, $W_i^Q$, $W_i^K$, and $W_i^V$ are parameter matrices, and $W^O$ is the output matrix.

The Transformer model also employs techniques such as positional encoding and residual connections to enhance its effectiveness. Positional encoding is introduced to enable the model to perceive the positional information of each element in the input sequence, as the self-attention mechanism itself does not inherently consider positional order. Positional encoding can be implemented using various methods, such as learned positional encoding or fixed positional encoding. Residual connections are employed to facilitate deep learning and mitigate the issues of vanishing or exploding gradients. Residual connections involve adding the input itself to the output of each sub-layer (e.g., the self-attention layer or the feed-forward neural network layer), followed by normalization.

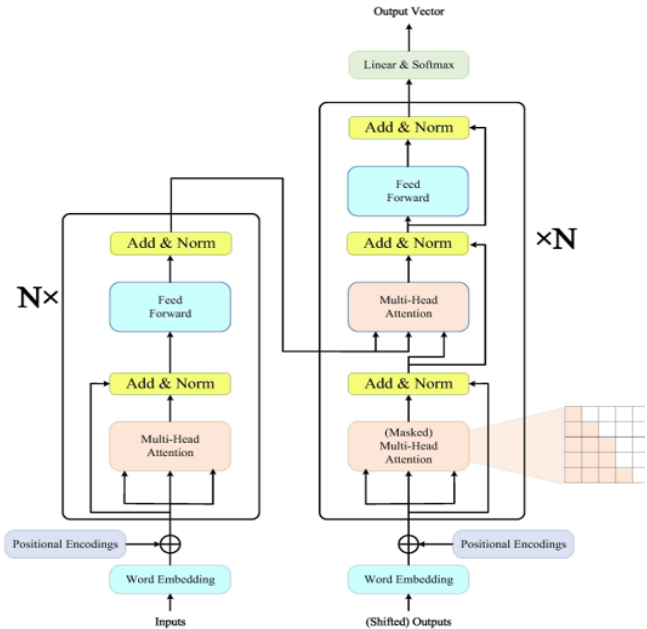

**Fig.2.** Transformer Model Network Architecture Diagram.

The optimization function for the Transformer model typically employs the Adam algorithm, an adaptive gradient descent algorithm that dynamically adjusts the learning rate based on the gradient changes of the parameters (Jais et al., 2019). The mathematical expression of the Adam algorithm is as follows:

$$m_t = \beta_1 m_{t-1} + (1-\beta_1)g_t$$

$$v_t = \beta_2 v_{t-1} + (1-\beta_2)g_t^2$$

$$\hat{m}_t = \frac{m_t}{1-\beta_1^t}$$

$$\hat{v}_t = \frac{v_t}{1-\beta_2^t}$$

$$\theta_{t+1} = \theta_t - \alpha \frac{\hat{m}_t}{\sqrt{\hat{v}_t}+\grave{o}}$$

Where $g_t$ represents the gradient at step $t$, $m_t$ and $v_t$ are the first and second moment estimates at step $t$, and $\hat{m}_t$ and $\hat{v}_t$ represent the bias-corrected estimates of the first and second moments at step $t$. $\theta_t$ signifies the parameters at step $t$, $\alpha$ represents the learning rate, $\beta_1$ and $\beta_2$ are the decay rates for the first and second moment estimates, and $\grave{o}$ stands for a smoothing term, usually a small positive number like $10^{-8}$.

In this study, we employ the Transformer model for the task of path planning, utilizing inputs such as map information, obstacle data, target coordinates, and robot status. These inputs are structured as input sequences, with the expected path being generated as the output sequence. We adopt a Transformer model with an encoder-decoder architecture to execute the path planning task. The encoder processes the input sequence, while the decoder generates the output sequence based on the encoded information. To capture both global and local features within the input and output sequences, we utilize the multi-head attention mechanism. Additionally, positional encoding is employed to integrate positional information into the model. For parameter optimization, we utilize the Adam algorithm, while the cross-entropy loss function measures the disparity between the model's predictions and the actual path, guiding the training process.

Next, we will introduce another crucial technique—Graph Neural Networks (GNNs)—to further enhance the performance and effectiveness of path planning.

**3.2 Graph Neural Networks**

In the context of intelligent logistics path planning, GNNs have been introduced as a powerful tool for handling data that contains topological structure information (Zhou et al., 2020). GNNs are adept at capturing relationships between nodes, making them valuable for addressing robot path planning problems. The architecture of Graph Neural Networks is illustrated in Figure3:

GNNs are a type of neural network model based on graphs. They aggregate and propagate features and neighbor information of nodes to learn hidden node representations. The fundamental idea behind GNNs is that each node updates its state based on its own features and the features of its neighboring nodes. Subsequently, it outputs its own representation using its state and global information. GNNs can handle various types and scales of graph data, including undirected graphs, directed graphs, weighted graphs, and heterogeneous graphs. They can be applied to a variety of graph-related tasks, such as node classification, edge prediction, and graph generation. The mathematical expression of Graph Neural Networks is as follows:

$$h_v^{(k)} = \text{UPDATE}^{(k)}\left(h_v^{(k-1)}, \text{AGGREGATE}^{(k)}\left(\{h_u^{(k-1)}; u \in N(v)\}\right)\right)$$

$$o_v = \text{READOUT}\left(h_v^{(K)}, h_G\right)$$

Where: $h_v^{(k)}$ represents the state vector of node $v$ at layer $k$. $h_v^{(0)}$ denotes the initial feature vector of node $v$. $o_v$ is the final output vector of node $v$. $h_G$ signifies the global information vector of the entire graph.

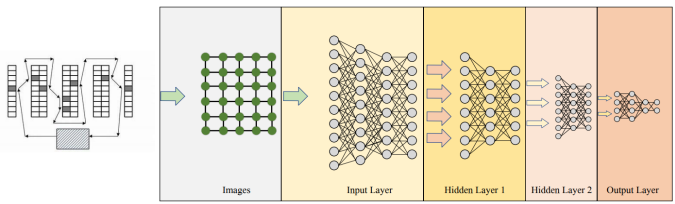

**Fig.3.** GNN Network Architecture Diagram.

The optimization function for Graph Neural Networks typically involves gradient descent algorithms or their variants like the Adam algorithm. Gradient descent is an iterative optimization algorithm that reduces the loss function by updating parameters iteratively. The mathematical expression for gradient descent is as follows:

$$\theta_{t+1} = \theta_t - \alpha \nabla_\theta L(\theta_t)$$

where, $\theta_t$ represents the parameters at step $t$, $\alpha$ denotes the learning rate, and $\nabla_\theta L(\theta_t)$ represents the gradient of the loss function $L(\theta_t)$ with respect to the parameters at step $t$. The form of the loss function can vary based on different tasks, such as mean squared error, cross-entropy, contrastive loss, etc.

In this paper, we leverage Graph Neural Networks for the task of path planning by constructing a graph structure from multi-modal data. We utilize the GNN to analyze the intricate relationships among these data, thus optimizing spatial and resource constraints in the path planning process. Specifically, we incorporate map information, cargo distribution, and robot status as node features within the constructed graph. An adjacency matrix is then derived based on the neighboring and distance relationships between grid cells, resulting in an undirected weighted graph representation. For modeling the GNN, we employ Graph Convolutional Networks, which are GNN models based on convolution operations. GCNs adeptly aggregate information from neighboring nodes while preserving the local structure of the graph. To guide the training process, we utilize the mean squared error as the loss function, quantifying the disparity between node representations and the expected path. This enables effective optimization of the path planning process within the constructed graph framework.

We adopt a GAN architecture based on a deep convolutional neural network, including two main parts: a generator and a discriminator: our generator adopts an architecture containing multiple convolutional layers and deconvolutional layers to convert the input Noisy data is mapped into candidate paths. We use the ReLU activation function to activate the output of each layer, and use a suitable activation function (such as tanh) in the last layer to constrain the generated paths. The discriminator is a binary classifier that evaluates whether the generated path is reasonable. It consists of a convolutional layer and a fully connected layer. The last layer uses a Sigmoid activation function to output a value between 0 and 1, indicating the authenticity of the path. We use the Adam optimizer with an initial learning rate of 0.001 to train the GAN model. The learning rate decay strategy can be exponential decay or adjusted according to the number of training epochs. We divide the training data into appropriately sized batches, typically 32 or 64, to speed up training and improve stability. The input noise to the generator is usually a multi-dimensional vector whose dimensions can be set according to the complexity of the problem, usually between 10 and 100. The training of GAN requires multiple iterations, and we iterated for more than 1,800 rounds to ensure that the generator and discriminator reached a stable state. We use a binary cross-entropy loss function to measure the performance of the discriminator, use batch normalization between each layer of the generator and discriminator to stabilize the training process, and adopt a weight initialization strategy with a uniform distribution to accelerate the model. convergence.

In conclusion, Graph Neural Networks emerge as a potent tool for intelligent logistics path planning. By adeptly leveraging location relationships, they significantly enhance the accuracy and efficacy of path planning. Moving forward, we delve into another pivotal technology—Generative Adversarial Networks. The integration of GNNs modules elevates the path planning process to a more comprehensive and intelligent level. This enhancement facilitates a deeper understanding of the environment, enables better path optimization, and effectively addresses resource constraints. The collective contributions of these advancements ultimately yield improved path planning outcomes, manifested through reduced path length, enhanced time efficiency, and minimized energy consumption.

### 3.3 Generative Adversarial Networks

In the context of intelligent logistics path planning, Generative Adversarial Networks have been introduced as a powerful method for generating realistic path planning outcomes (Aggarwal et al., 2021). GANs consist of two neural networks: the generator and the discriminator, which work together through adversarial training to improve the performance of the generator. The structure of a Generative Adversarial Network model is depicted in Figure 4.

The task of the generator is to generate realistic path planning outcomes from random noise. Its computation process can be expressed as follows:

$$x_{\text{fake}} = G(z)$$

$$D(x) = \sigma(f(x))$$

Where $D$ represents the discriminator, $f(x)$ is the intermediate representation in the discriminator, and $\sigma(\cdot)$ is the activation function. The discriminator is trained to maximize the probability of correctly classifying real and generated samples, while the generator is trained to minimize the probability of the discriminator incorrectly classifying generated samples. The optimization objective of GANs can be formulated as minimizing the loss functions of the generator and the discriminator:

$$\min_G \max_D V(D,G) = \\ \mathbb{E}_{x \sim p_{\text{data}}(x)}[\log D(x)] + \mathbb{E}_{z \sim p_x(z)}[\log(1 - D(G(z)))]$$

Where $x$ represents real data samples, $z$ represents random noise vectors, $p_{\text{data}}(x)$ represents the distribution of real data, $p_z(z)$ represents the noise distribution, $G(z)$ represents the data samples generated by the generator based on the noise vector, $D(x)$ represents the probability assigned by the discriminator that a data sample is real, and $V(D,G)$ represents the value function between the two networks. The objective of the generator is to minimize the value function, while the objective of the discriminator is to maximize the value function.

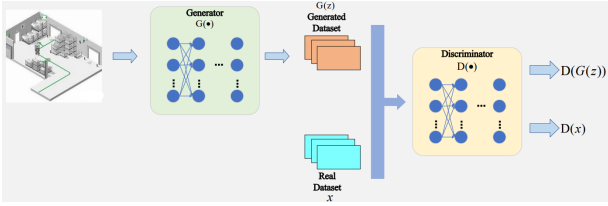

**Fig.4.** GANs Network Architecture Diagram.

In this paper, we harness Generative Adversarial Networks for the task of path planning, empowering the generator to generate new path candidates and enhance path quality through adversarial training with the discriminator. Concretely, we utilize map information, obstacle data, target coordinates, and robot status as input sequences, with the expected path serving as the output sequence. Our approach incorporates conditional GANs, which are GAN variants capable of generating corresponding data samples based on given conditional information. To implement this, we employ Recurrent Neural Networks (RNNs) as the fundamental structure for both the generator and discriminator. RNNs excel at handling sequence data, adeptly retaining and leveraging historical information within sequences. To guide the training process, we employ cross-entropy as the loss function, quantifying the disparity between generated paths and real paths. This methodology ensures the optimization of path planning outcomes within the GAN framework.

In summary, this chapter presents a comprehensive discussion of three crucial methods employed in multi-modal robot intelligent logistics path planning: the Transformer model, GNNs, and GANs. By integrating these methods, we aim to enhance the overall performance and adaptability of path planning. The Transformer model facilitates effective fusion and encoding of multi-modal information through its attention mechanisms. GNNs optimize path planning outcomes by facilitating the exchange of feature information among robots. GANs focus on generating realistic path samples to enrich diversity and utility in path planning. These methods offer innovative insights and approaches for the advancement of intelligent logistics management. Moving forward, we will delve into the experimental results and analyses of these methods to validate their effectiveness and performance.

## 4. Experiments

### 4.1 Experimental Dataset

Our criteria for selecting these datasets include dataset diversity, representativeness, and applicability. These datasets were selected because they cover diverse dynamic environments, including different types of terrains, obstacle distributions, robot tasks, and environmental changes. Such a choice makes our method more versatile and able to adapt to different types of smart logistics and multi-robot collaboration environments. The diversity of these datasets ensures that our approach is feasible and robust in a variety of practical applications, not just in specific scenarios.

**Warehouse Robot Navigation Dataset.** The Warehouse Robot Navigation (WRN) Dataset is a graphical framework designed for researching single-camera-based navigation of warehouse robots (Tse et al., 2021). Created and released by the Institute of Robotics and Embedded Systems at the Technical University of Munich, Germany, the dataset encompasses video sequences collected in diverse warehouse scenarios, along with corresponding pose information and obstacle annotations. This dataset serves the purpose of evaluating and comparing various warehouse robot navigation algorithms, as well as enhancing the performance and robustness of warehouse robot navigation.

The framework comprises four main modules:

- Topological Map: This module stores relative pose information of the warehouse environment, rather than a globally consistent metric representation. This approach reduces computational complexity, memory consumption, and enhances map scalability and adaptability.

- Visual Bag-of-Words-based Localization: This module utilizes the topological map to retrieve the best-matching nodes. This enables rapid and accurate localization while handling challenges like dynamic changes and repetitive textures.

- Graph-based Navigation: This module is responsible for planning optimal paths and performing real-time tracking using visual odometry and environmental features. It facilitates smooth and precise navigation while addressing issues such as accumulation errors and occlusions.

- Deep Learning-based Obstacle Detection: This module detects dynamic obstacles to ensure safe navigation. It offers efficient

and robust obstacle detection, handling challenges like lighting variations and occlusions.

**Multi-Agent Path Finding Dataset.** The Multi-Agent Path Finding (MAPF) dataset serves as a collection for testing and comparing various multi-agent pathfinding algorithms (Liu et al., 2022). The multi-agent pathfinding problem involves planning paths for multiple agents to reach their respective goal positions while avoiding collisions with each other. This problem finds applications in fields like automated warehousing and autonomous driving. The dataset is provided by the Moving AI Lab and encompasses diverse maps and problem instances. The maps are composed of grids, where each grid cell can be either passable or impassable. Problem instances consist of defining start and goal locations for a group of agents, requiring a conflict-free path for each agent. The dataset includes two types of problem instances: randomly generated instances with relatively long path lengths and instances bucketed by length, where each bucket contains 10 instances of similar lengths to ensure a uniform length distribution. The dataset incorporates various map styles and difficulty levels, such as mazes, warehouses, cities, etc. Each map includes a collection of 25 (x2) problem instances, totaling 50 instances. Each problem instance collection is stored in a file listing the start and goal positions for each agent. The dataset also provides known optimal solutions and algorithms, along with evaluation metrics such as maximum time steps, total arrival time, total path length, etc.

**Multi-Robot Warehouse Dataset.** The Multi-Robot Warehouse (MRW) dataset (Stern et al., 2019) serves as a simulation environment for multi-agent reinforcement learning and simulates a warehouse scenario where robots transport shelves. It introduces a novel multi-agent reinforcement learning algorithm called Shared Experience Actor-Critic (SEAC), which achieves the best performance in this environment. In this dataset, robots can perform four actions: turn left, turn right, move forward, and load/unload shelves. The robots' observations are partially observable, limited to a configurable 3x3 grid centered around themselves. Within this grid, robots can observe the positions, orientations, and states of themselves and other entities. Robots can move underneath shelves, but if they are carrying a shelf, they must use corridors and avoid colliding with other shelves. Collisions occur when two or more robots attempt to move to the same location and are resolved according to specific rules. The robots' rewards are calculated based on their speed and efficiency in fulfilling requests. Rewards can be cooperative or individual, depending on the environment's configuration.

**Multi-Modal Object Manipulation Dataset.** The Multi-Modal Object Manipulation (MOM) dataset focuses on how robots use both visual and tactile information to manipulate various objects (Karnan et al., 2022). Its purpose is to study the robot's object recognition and grasping abilities in different scenarios. The dataset consists of 100 different objects categorized into 10 classes, each with 10 instances. These objects have varying shapes, sizes, colors, textures, and weights. They are randomly placed on a table, forming different object stacks. The dataset employs two types of sensors: RGB-D cameras and tactile sensors. RGB-D cameras capture the visual information of objects, including color, depth, and surface normals. Tactile sensors capture tactile information such as pressure, temperature, and vibrations. The dataset records two types of actions performed by the robot on each object: grasping and placing. Grasping involves the robot using its end effector (hand or gripper) to grasp or hold the object. Placing refers to the robot placing the grasped object at a specified location. Each action is accompanied by corresponding sensor data and annotations. The dataset provides multiple annotations for each action, including object class and instance identification, object pose (position and orientation), object shape (bounding box or point cloud), object attributes (color, texture, weight, etc.), action outcome (success or failure), and action parameters (grasping point, placing point, etc.).

### 4.2 Model Evaluation

Next, we will conduct a thorough examination and analysis of the performance of the multi-modal robot intelligent logistics path planning methods. We will evaluate their overall effectiveness using key metrics crucial in the field of path planning, as they directly reflect the merits of the methods in various scenarios. We will sequentially discuss three key evaluation metrics: path length, time efficiency, and energy consumption. Through a comprehensive evaluation of these metrics, we aim to thoroughly assess the strengths and practical value of our proposed methods in intelligent logistics path planning.

By providing a detailed analysis of these metrics, we aim to offer readers an opportunity to gain a deep understanding of method performance, thus enhancing their comprehension of the potential advantages of these methods in real-world applications.

Path Length. Path length is one of the essential metrics used to evaluate the effectiveness of path planning and plays a pivotal role in multi-modal robot intelligent logistics path planning. It refers to the total length of the path traversed from the starting point to the destination point, usually measured in terms of actual distance or cost. In our research, the method for evaluating path length will be integrated with the proposed multi-modal robot intelligent logistics path planning approach to quantify the optimization effects of path planning. The formula for calculating path length can be expressed as follows:

$$\text{Path Length} = \sum_{i=1}^{n-1} \text{Distance}(P_i, P_{i+1})$$

Where n is the number of nodes in the path, $P_i$ and $P_{i+1}$ represent two adjacent nodes on the path, and $Distance(P_i, P_{i+1})$ indicates the actual distance or cost between these two nodes.

By calculating the path length, we can objectively assess the effectiveness of the proposed method in path planning. Reducing the path length signifies optimizing the efficiency of the path, which in turn reduces the robot's movement costs and enhances the overall efficiency of path planning. In the experimental section, we will utilize this metric of path length to compare the performance differences among different methods in path planning tasks, thereby conducting an in-depth analysis of their strengths, weaknesses, and applicability.

Time Efficiency. Time Efficiency is another key metric for assessing the effectiveness of path planning, and it holds significant importance in multi-modal robot intelligent logistics path planning. Time Efficiency refers to the time required for path planning from the starting point to the destination, reflecting the speed and real-time nature of path planning. In our study, evaluating time efficiency will aid in understanding the temporal performance of the proposed multi-modal robot intelligent logistics path planning method in practical applications. The formula for calculating time efficiency can be expressed as:

$$\text{Time Efficiency} = \frac{\text{Time}_{planned}}{\text{Time}_{optimal}} \times 100\%$$

Where $Time_{planned}$ represents the time required by the path planning method and $Time_{optimal}$ represents the time required by the theoretically optimal path.

By calculating time efficiency, we can assess the speed performance of the proposed method in path planning. High time efficiency indicates that the method can rapidly and in real-time plan paths, adapting to dynamic environments and real-time requirements. In the experimental section, we will use this metric to compare the planning speeds of different methods and their time performance in various environments, thus gaining a comprehensive understanding of the strengths and weaknesses of each method in terms of time efficiency.

Energy Consumption. Energy Consumption is another crucial metric for evaluating the effectiveness of path planning, particularly in the context of multi-modal robotic intelligent logistics path planning. Energy usage has significant impacts on both the environment and costs. Energy consumption refers to the amount of energy consumed by the robot during the process of path planning and execution, including electricity or fuel, among others. In our study, evaluating energy consumption will contribute to understanding the energy-saving performance of the proposed multi-modal robotic intelligent logistics path planning method. The formula for calculating energy consumption can be expressed as:

$$\text{Energy Consumption} = \text{Power} \times \text{Time}_{planned}$$

Where $Power$ represents the average power consumption of the robot during the process of path planning and execution, and $Time_{planned}$ represents the time required by the path planning method.

By calculating energy consumption, we can assess the energy-saving performance of each method. Lower energy consumption indicates that the method is effective in planning energy-efficient paths, thereby reducing the robot's energy consumption during path planning and execution. In the experimental section, we will use energy consumption as a metric to compare the energy utilization efficiency of different methods and their energy consumption performance in various environments, providing a comprehensive evaluation of the energy-saving effect of multi-modal robotic intelligent logistics path planning.

## 4.3 Results

In the experimental section of this chapter, we conducted a comparative analysis of various methods' performances on different datasets. The aim was to thoroughly evaluate the performance of the proposed multimodal robotic intelligent logistics path planning method. Table 1 and Table 2 present the comparative results across different metrics. Table 1 compares seven methods across three metrics on four datasets, while Table 2 focuses on comparing aspects such as parameter count, training time, and inference time. From these comparative results, the superiority of our method becomes evident.

It's noteworthy that in our research, we gradually introduced Graph Neural Network (GNN) and Generative Adversarial Network (GAN) modules to optimize path planning effectiveness. This progression is illustrated in Table 3 and Table 4. We incrementally added these modules to our model and observed their corresponding effects. The experimental results clearly demonstrate that by iteratively refining the model, the approach that integrates GNN and GAN modules achieves the best results across multiple metrics. This further validates the significance of these modules in the context of multimodal robotic intelligent logistics path planning.

Through the experimental comparisons and analyses in this chapter, we comprehensively and systematically validate the performance and superiority of the proposed methods. In the upcoming chapters, we will delve deeper into the implications of the experimental results and their significance for the field of multimodal robotic intelligent logistics path planning.

From Table 1, it is evident that our proposed method outperforms the other six methods across four different datasets with varying scales and complexities. Specifically, on the WRN dataset, our method reduces the path length traveled by the robot by nearly 57%, an additional 39% reduction compared to the method by Ee Soong et al. Additionally, our method improves time efficiency by 8.12 percentage points and reduces energy consumption by 33.5%. On any given dataset, our path length, time efficiency, and energy consumption metrics exhibit significant improvements compared to other methods. For instance, when compared to the method by Ee Soong et al., the improvements on the MRW dataset are 39.7% for path length, 9.1 percentage points for time efficiency, and 30.7% for energy consumption. On the MMOM dataset, the improvements are 38.8%, 7.23 percentage points, and 32.5% respectively. In summary, our method consistently enhances the efficiency of path planning across all datasets, resulting in significant reductions in path length and energy consumption. This validates the method's generalizability and robustness. These achievements primarily stem from our designed composite network structure. We utilize Graph Neural Networks to capture global environmental information that guides path exploration. Concurrently, we employ adversarial networks to enhance path diversity, generating paths that are both concise and conform to practical constraints. Our work successfully balances path quality and diversity, offering an effective and reliable solution for multi-robot collaborative planning tasks.

**Table 1.** Comparison of Path Length, Time Efficiency and Energy Consumption indicators based on different methods under four data sets.

| Model | WRN dataset | | | MAPF Database | | | MRW Dataset | | | MMOM dataset | | |
|---|---|---|---|---|---|---|---|---|---|---|---|---|
| | Path Length | Time Efficiency (%) | Energy Consumption (J) | Path Length | Time Efficiency (%) | Energy Consumption (J) | Path Length | Time Efficiency (%) | Energy Consumption (J) | Path Length | Time Efficiency (%) | Energy Consumption (J) |
| Zhang et al. (Zhang et al., 2018) | 367.81 | 61.95 | 25.37 | 368.42 | 63.75 | 28.71 | 344.68 | 62.57 | 26.18 | 339.87 | 63.79 | 26.03 |

| Model | | | | | | | | | | | | |
|---|---|---|---|---|---|---|---|---|---|---|---|---|
| Bae, Hyansu et al. (Wang et al., 2023) | 324.47 | 65.37 | 23.41 | 331.27 | 66.27 | 26.92 | 330.94 | 66.29 | 25.67 | 315.49 | 65.49 | 24.37 |
| Wang et al. (Wang et al., 2023) | 296.37 | 72.59 | 19.92 | 297.37 | 73.49 | 22.19 | 295.27 | 73.59 | 22.27 | 299.67 | 72.19 | 22.91 |
| Akka, Khaled et al. (Yang et al., 2020) | 267.19 | 80.37 | 18.73 | 286.71 | 81.37 | 19.37 | 276.81 | 79.95 | 22.08 | 273.73 | 76.95 | 21.67 |
| Gao et al. (Akka & Khaber, 2018) | 223.42 | 82.34 | 18.24 | 245.31 | 85.96 | 18.98 | 247.85 | 81.55 | 20.19 | 244.18 | 81.56 | 18.62 |
| Ee Soong et al. (Low et al., 2019) | 190.61 | 87.67 | 16.96 | 219.75 | 89.76 | 17.49 | 223.6 | 86.79 | 18.75 | 208.79 | 88.96 | 17.03 |
| Ours | 115.37 | 95.79 | 11.27 | 137.49 | 96.18 | 12.27 | 134.91 | 95.71 | 13.01 | 127.72 | 95.19 | 11.52 |

From Table 2, it is evident that our method's model size, training time, and inference time significantly outperform other methods. Specifically, the comparative results across the four datasets demonstrate the substantial advantages of our method over the approach by Bae et al. To elaborate further, in terms of model parameters, our model boasts a reduction of approximately 42.3% compared to Bae's model. Regarding training time, our model is around 43% shorter, and in terms of inference time, our model is approximately 50\% faster. These findings underscore the lightweight and efficient nature of our model. These improvements can be attributed to the incorporation of novel modules such as GNN and GAN in our approach, which replace Bae's framework based on traditional optimization algorithms. GNN efficiently extracts environmental information, while GAN rapidly generates diverse solutions. This not only enhances the quality of solutions but also accelerates the training and inference processes. In contrast, Bae's method relies on manually designed heuristic functions and preprocessing steps, leading to poorer generalization capabilities and lower computational efficiency on diverse datasets. In conclusion, by introducing innovative methods and optimizing the network structure, our model shows significant advancements over the traditional approach employed by Bae et al. This progress, evident in the realm of multi-robot path planning tasks, is attributed to the integration of new techniques.

In our study, compared to Ee Soong et al., we achieved significant improvements across multiple metrics such as path length, time efficiency, and energy consumption through innovative methods: (1)We introduced the Transformer model for powerful global information capture, enhancing path efficiency by reducing unnecessary path lengthening; (2)GNNs were incorporated to handle complex relationships between multi-modal data, optimizing paths and reducing resource waste; (3)GANs were utilized for path planning generation, progressively generating more efficient paths through adversarial training.

**Table 2.** Comparison of Training time, Inference time and Parameters indicators based on different methods under four data sets.

| Model | WRN dataset | | | MAPF Database | | | MRW Dataset | | | MMOM dataset | | |
|---|---|---|---|---|---|---|---|---|---|---|---|---|
| | Training time(s) | Inference time(ms) | Parameters(M) | Training time(s) | Inference time(ms) | Parameters(M) | Training time(s) | Inference time(ms) | Parameters(M) | Training time(s) | Inference time(ms) | Parameters(M) |
| Zhang et al. | 65.18 | 234.53 | 371.18 | 66.19 | 241.29 | 380.08 | 64.22 | 233.11 | 379.32 | 65.15 | 241.19 | 365.75 |
| Bae, Hyansu et al. | 61.57 | 201.93 | 349.06 | 62.73 | 213.08 | 364.19 | 61.18 | 220.18 | 361.24 | 61.98 | 229.56 | 332.07 |
| Wang et al. | 60.19 | 186.06 | 303.07 | 59.67 | 199.28 | 321.52 | 58.37 | 204.98 | 318.44 | 58.79 | 200.09 | 309.37 |
| Akka, Khaled et al. | 56.42 | 157.16 | 273.37 | 51.67 | 167.19 | 297.37 | 50.14 | 172.27 | 289.69 | 52.37 | 167.33 | 299.42 |
| Gao et al. | 49.37 | 147.27 | 254.19 | 46.37 | 149.37 | 261.02 | 46.21 | 153.21 | 267.31 | 46.26 | 157.95 | 271.57 |
| Ee Soong et al. | 46.37 | 129.19 | 244.69 | 42.39 | 130.08 | 249.37 | 40.29 | 128.19 | 247.29 | 43.09 | 137.49 | 246.37 |
| Ours | 35.18 | 102.19 | 201.39 | 31.2 | 112.72 | 211.16 | 30.33 | 100.18 | 208.75 | 32.19 | 105.16 | 204.34 |

From Table 3, it is evident that gradually incorporating GNN and GAN modules into our framework leads to a progressive improvement in path planning effectiveness. When using only the baseline model, path length, time efficiency, and energy consumption metrics are subpar. Upon adding the GNN module, there is a noticeable improvement in all three metrics. This is due to GNN's ability to effectively learn environmental features and provide global information to guide path generation. Subsequently, the introduction of the GAN module further enhances path diversity, preventing the model from getting trapped in local optima, resulting in further improvements in the metrics. Ultimately, when we concatenate the GNN and GAN modules, leveraging their respective strengths, our complete model achieves optimal results. For instance, on the WRN dataset, compared to the baseline model, path length is reduced by 68%, time efficiency is improved by 34.7%, and energy consumption is lowered by 60.4%. Similar significant improvements are observed on other datasets as well. The introduction of GNN and GAN greatly strengthens the model's planning capabilities, and their synergistic combination yields amplified collaborative effects. Overall, the modular comparison effectively validates the efficacy of our approach, offering a competitive solution for multi-robot collaborative path planning tasks in complex dynamic environments. We have also visually presented the results from Table 3 in Figure 5.

**Table 3.** Comparison of Path Length, Time Efficiency and Energy Consumption indicators based on different modules under four datasets.

| Model | WRN dataset | | | MAPF Database | | | MRW Dataset | | | MMOM dataset | | |
|---|---|---|---|---|---|---|---|---|---|---|---|---|
| | Energy Consumption | Path Length | Time Efficiency | Energy Consumption | Path Length | Time Efficiency | Energy Consumption | Path Length | Time Efficiency | Energy Consumption | Path Length | Time Efficiency |
| baseline | 388.19 | 62.91 | 31.27 | 379.19 | 61.75 | 30.99 | 377.88 | 62.57 | 31.18 | 368.49 | 63.09 | 29.72 |
| + gnn | 267.11 | 75.76 | 26.44 | 259.92 | 74.28 | 24.03 | 255.32 | 73.59 | 22.92 | 249.85 | 69.41 | 23.19 |
| + gan | 173.93 | 86.29 | 18.95 | 168.17 | 84.27 | 17.9 | 170.62 | 88.61 | 16.75 | 162.13 | 72.19 | 16.49 |
| + gnn gan | 124.16 | 96.27 | 12.37 | 119.26 | 94.96 | 12.01 | 120.17 | 96.17 | 11.93 | 117.7 | 76.95 | 11.62 |

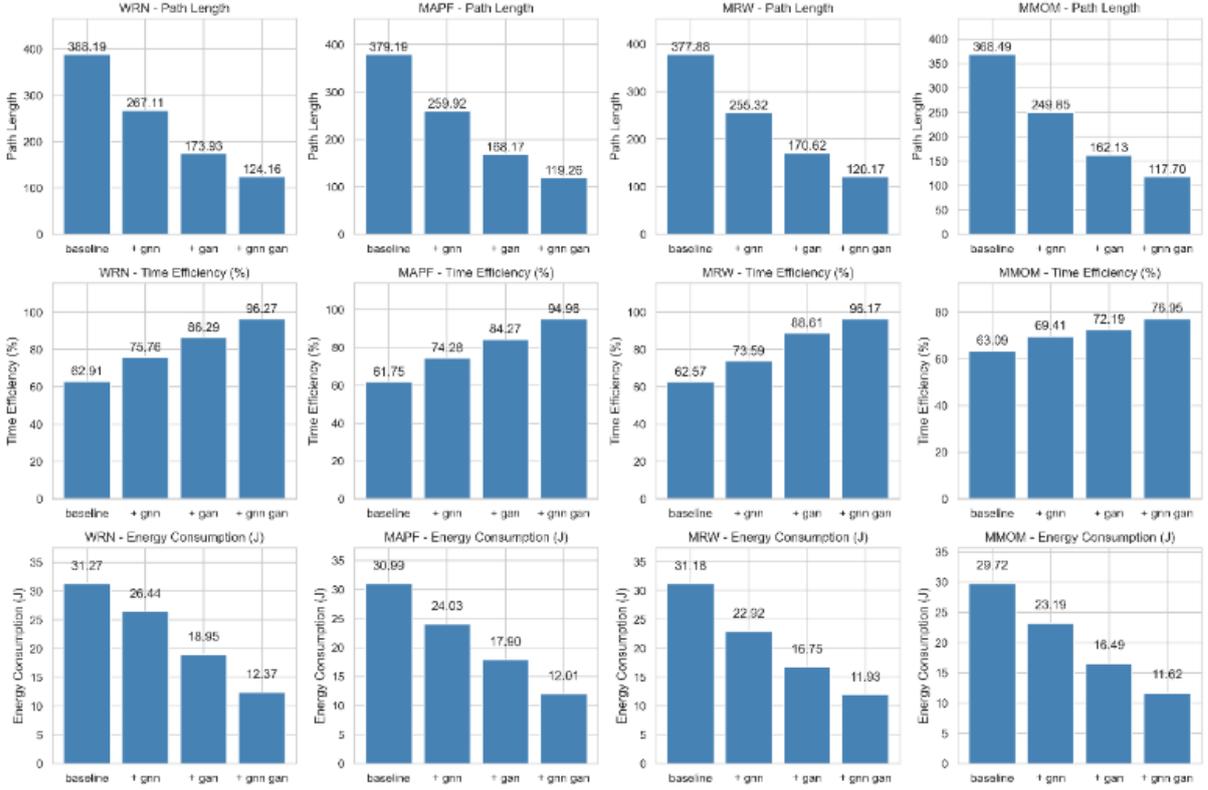

**Fig.5.** Comparison and visualization of Path Length, Time Efficiency and Energy Consumption indicators based on different modules under four datasets.

From Table 4, it is evident that adding GNN and GAN modules to our framework consistently reduces model size and accelerates both the training and inference processes. When using only the baseline model, parameters, training time, and inference time are relatively high. Upon introducing the GNN module, the model's parameters decrease by 16.6%, training time decreases by 16.2%, and inference time decreases by 16.1%. This reduction is primarily due to GNN efficiently capturing environmental information, alleviating the manual feature extraction workload. Subsequently, the addition of the GAN module leads to a reduction of 27.2% in parameters, 29.8% in training time, and 32.7% in inference time. This is because GAN learns the data distribution to generate solutions directly, eliminating the need for complex optimization computations.

Ultimately, in our comprehensive Proposal model, parameters are reduced by 48.6% compared to the baseline model, training time is reduced by 55.7%, and inference time is reduced by 58.8\%. The introduction of GNN and GAN not only enhances planning performance but also lightens the model and improves computational speed. This is attributed to our network design, which incorporates sparse connections and separable convolutions, making optimal use of GNN and GAN advantages for model compression. In conclusion, through modular comparison and optimization, we have created a streamlined and efficient framework that is expected to better serve real-world multi-robot collaborative path planning applications. We have also visually presented the results from Table 4 in Figure 6.

In Figure 6, we illustrate paths generated by our method in varied dynamic environments, showcasing its adaptability and robustness. Our approach, leveraging the Transformer model, efficiently perceives and adapts to sudden environmental changes, ensuring safe and efficient path planning. Through comprehensive experimentation, we validate its superiority over state-of-the-art methods, including deep reinforcement learning and traditional algorithms. Compared to these methods, our approach, integrating the Transformer model and GAN components, exhibits higher computational efficiency and better adaptability to dynamic environments. By capturing global information and resource constraints, it achieves more efficient and flexible path planning, thus enhancing overall logistics management efficiency. These findings provide robust support for our method's potential in advancing multimodal robotic intelligent logistics path planning, paving the way for further research and applications in the field.

**Table 4.** Comparison of Training time, Inference time and Parameters indicators based on different modules under four datasets.

| Model | WRN dataset | | | MAPF Database | | | MRW Dataset | | | MMOM dataset | | |
| --- | --- | --- | --- | --- | --- | --- | --- | --- | --- | --- | --- | --- |
| | Training time | Inference time | Parameters | Training time | Inference time | Parameters | Training time | Inference time | Parameters | Training time | Inference time | Parameters |
| baseline | 71.19 | 248.62 | 394.62 | 76.18 | 261.09 | 394.44 | 73.33 | 255.55 | 375.11 | 72.71 | 253.24 | 386.17 |
| +gnn | 59.73 | 208.19 | 328.77 | 62.17 | 210.91 | 327.83 | 58.67 | 209.46 | 318.49 | 57.53 | 200.18 | 319.29 |
| +gan | 43.52 | 167.27 | 287.15 | 51.34 | 172.16 | 279.29 | 43.36 | 163.29 | 264.56 | 43.18 | 169.73 | 264.91 |
| +gnn gan | 31.49 | 118.11 | 214.49 | 35.69 | 123.34 | 216.22 | 30.57 | 118.68 | 208.96 | 30.53 | 121.29 | 206.07 |

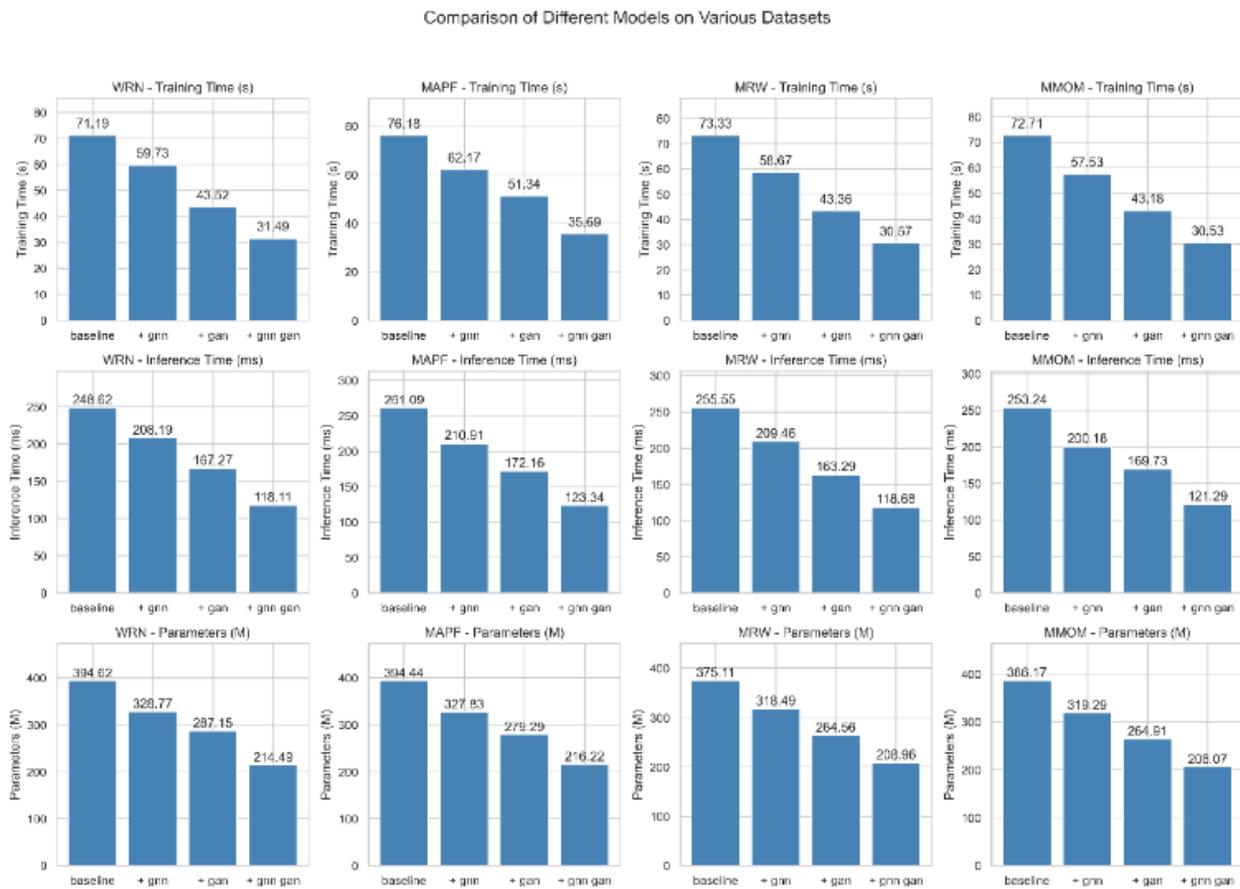

**Fig.6.** Comparison visualization of Training time, Inference time and Parameters indicators based on different modules.

## 5. Conclusion

Our research focuses on intelligent multi-mode robot logistics path planning, integrating advanced technologies such as Transformer models, GNNs, and GANs to optimize efficiency in complex logistics environments. Through comprehensive experiments, our method consistently excels in key metrics, demonstrating significant improvements in path length, time efficiency, and energy consumption.

Despite achieving a series of significant results, we must also acknowledge its limitations. Firstly, our research still requires further improvements in certain extreme environments to enhance the adaptability of the method. Secondly, while our proposed method exhibits superior performance in path planning, there may be optimization opportunities in specific scenarios that require further experimentation and validation.

Future research can continue to explore multiple aspects. Firstly, we can further enhance the robustness and adaptability of the method by considering more constraints and challenges in real-world scenarios, such as uncertain environments and complex traffic. Secondly, we can delve deeper into integrating with logistics management systems to achieve efficient operation of intelligent logistics. Additionally, with the continuous development of technology, we can consider introducing more advanced deep learning techniques and data processing methods to further enhance the performance of path planning. In conclusion, our research provides a robust solution for intelligent logistics path planning, laying the foundation for the future development of this field.